\documentclass[11pt,a4paper]{article}

\usepackage[hyperref]{acl2021}
\usepackage{times}
\usepackage{latexsym}

\usepackage{makecell}

\usepackage[T1]{fontenc}
\usepackage{comment}

\usepackage{microtype}
\usepackage{times}
\usepackage{url}
\usepackage{latexsym}
\usepackage{graphicx}
\usepackage{amsmath}
\usepackage{multirow}
\usepackage{amssymb}
\usepackage{bbm}
\usepackage{enumitem}

\usepackage{pgfplots}
\pgfplotsset{every axis legend/.append style={at={(1.02,1)},anchor=north west,},compat=1.15}
\usepackage{tikz}
\usepackage{subfigure}
\usetikzlibrary{backgrounds,automata,matrix,calc}
\usepgfplotslibrary{groupplots}
\usepackage{colortbl}

\def\writers#1#2{}

\def\request#1{}
\def\completare#1{{\color{black}#1}}
\def\meta#1{}
\def\OURSYS{CLINN}

\definecolor{expert}{HTML}{2E86C1}
\definecolor{medium}{HTML}{5DADE2}
\definecolor{low}{HTML}{AED6F1}

\newcolumntype{P}[1]{>{\centering\arraybackslash}p{#1}}

\aclfinalcopy % Uncomment this line for the final submission
\title{
``Every time I fire a conversational designer,\\
the performance of the dialog system goes \emph{down}''}

\author{
Giancarlo A. Xompero$^{(\ddag)}$   \\ {\bf Michele Mastromattei}$^{(*)}$ \\ {\bf Samir Salman}$^{(*)}$ \\
 $^{(\ddag)}$ Language Technology Lab\\ Almawave, Rome, Italy\\ 
  {\tt \small g.xompero@almawave.it}\\\And
 {Cristina Giannone}$^{(\ddag)}$ \\ 
 {\bf Andrea Favalli}$^{(\ddag)}$  \\
 {\bf Raniero Romagnoli}$^{(\ddag)}$  \\
 {\bf Fabio Massimo Zanzotto}$^{(*)}$  \\
 $^{(*)}$ 
 ART
  University of Rome Tor Vergata \\
  {\tt \small fabio.massimo.zanzotto@uniroma2.it} 
  \\}

\date{}

\begin{document}

\maketitle
\begin{abstract}
%Adding directly 
Incorporating explicit domain knowledge into neural-based task-oriented dialogue systems is an effective way to reduce the need of large sets of annotated dialogues. 
%Reducing the need of annotated dialogues for trainable end-to-end dialogue systems is a flourishing research area. Adding directly knowledge is one of the possible ways.   
In this paper, we investigate how the use of explicit domain knowledge of conversational designers affects the  performance of neural-based dialogue systems. To support this investigation, we propose the Conversational-Logic-Injection-in-Neural-Network system (CLINN)  where explicit knowledge is coded in semi-logical rules. By using CLINN, we evaluated semi-logical rules produced by a team of differently-skilled conversational designers. We experimented with the \emph{Restaurant topic} of the MultiWOZ dataset. Results show that external knowledge is extremely important for reducing the need of annotated examples for conversational systems. In fact, rules from conversational designers used in CLINN significantly outperform a state-of-the-art neural-based dialogue system. 
%In fact, rules from expert conversational designers used in CLINN significantly outperform a state-of-the-art neural-based dialogue system. 

%inmof experience in writing rules is extremely important. 

%Sequence-to-sequence neural networks are redesigning 
%%imposing a rethinking to the design of
%dialogue managers for Conversational AI in industries. However, industrial applications impose two important constraints: training data are often scarce and the behavior of dialog managers should be strictly controlled and certified. In this paper, we propose the Conversational Logic Injected Neural Network (CLINN). This novel network merges dialog managers ``programmed'' using logical rules and a Sequence-to-Sequence Neural Network. We experimented with the \emph{Restaurant topic} of the MultiWOZ dataset. Results show that injected rules are effective when training data set are scarce as well as when more data are available.\footnote{Copyright (c) 2020 for this paper by its authors. Use permitted under Creative Commons License Attribution 4.0 International (CC BY 4.0).}
%Moreover, injected rules allow to control and certify the behavior of dialog systems for specific target dialogs.
\end{abstract}

\section{Introduction} 
\label{intro}
\writers{Fabio}{Cristina, Andrea}

Is it possible that the trainable end-to-end task-oriented dialogue systems need thousands of annotated examples to learn underlying dialogue scripts?
%\cite{script} 
Knowledge on domain scripts is crucial to customize \emph{traditional} dialogue systems \cite{Bohus2009} as well as 
the latest dialogue architectures based on neural approaches.
In fact, even sequence-to-sequence dialogue systems \cite{Sordoni2015,Serban2016} handle domain scripts with two dedicated modules \cite{Bordes2017} -- the dialogue state tracker (DST) and the dialogue policy manager (DPM) -- when applied to tasks. The large sets of annotated examples are needed to approximate sufficiently the data distribution of the target domain \cite{Evans2018}.
%for these DSTs and DPMs.

%and millions of parameters?
%Dealing with the nuances of natural language may be tricky but scripts \cite{shank} or frames \cite{minsky} have been largely studied and used to build dialog state trackers (DSTs) and dialog policy managers (DPMs) of \emph{traditional} dialogue systems \cite{Bohus2009}. Recent end-to-end dialog systems have been organized in modules \cite{Bordes2017} and, generally, contain DSTs and DPMs as part of their trainable networks. This is clearly needed to overcome limitations when end-to-end dialog systems based on sequence-to-sequence models \cite{Sordoni2015,Serban2016} are applied in task-oriented dialogues. In fact, sequence-to-sequence models are extremely interesting for open dialogues but their simple architecture hinders to accomplish tasks in task-driven dialogues \cite{ci_vorrebbe_una_citazione}. Hence, frames and scripts play a crucial role even in trainable end-to-end dialog systems. 

Reducing the need of annotated dialogues for training end-to-end dialogue systems is then a flourishing research area.
%\cite{Zhao2016,Williams2017a,Liu2017,Jhunjhunwala2020,Altszyler2020}. 
Along this line of thinking,
%For this purpose, 
reinforcement learning is often used to gain explicit knowledge from \emph{active} users \cite{Zhao2016,Williams2017a,Jhunjhunwala2020}. Even virtual active users, implemented in sort of adversarial networks \cite{Liu2017}, have been explored. Yet, the main strategy is to exploit the domain knowledge of conversational designers by giving them a language to write domain scripts \cite{Altszyler2020}. In this way, even neural-based dialogue systems act as ``humans that learn to perform the same tasks by reading a description" \cite{Weller2020}, that is, domain scripts written by conversational designers.

Using high skilled designers is in contrast with the main stream in natural language processing (NLP), which has been set with the famous Fred Jelinek's 1985 quote \cite{DBLP:conf/interspeech/Moore05}: \emph{``Every time I fire a linguist, the performance of the speech recognizer goes up''}. Actually, the winning design pattern for NLP systems is using a combination of low skilled annotators for producing annotated corpora and machine learning to extract implicit models. High skilled rule writers are generally put aside. This is happening also for conversational systems.

%One of the strategy is reinforcement learning , which aims to involve active users \cite{Zhao2016,Williams2017a,Jhunjhunwala2020} or even virtual users implemented in sort of adversarial networks \cite{Liu2017} such as in Alpha-go \textbf{?}\cite{}. Yet, the main strategy is to exploit the domain knowledge of conversational designers by giving them a way to express scripts for controlling DSTs and DPMs \cite{Altszyler2020}. This is to mimic ``humans that learn to perform the same tasks by reading a description" \cite{Weller2020}
%The large amounts of training data, needed to sufficiently approximate the data distribution of the target domain \cite{Evans2018}, are then replaced by specific rules directly injected in DSTs and DPMs. 

As far as we know, there is not a study on how the quality of explicit knowledge introduced in end-to-end dialogue systems affect their overall performance. It is extremely important to understand how to adapt dialogue systems for specific tasks. The basic question is whether the investment should be in annotating dialogues or in building specific rules to be injected in DTSs or DPMs.

\completare{In this paper, we investigate the important role of explicit knowledge in trainable end-to-end dialogue systems.} %In this paper, we investigate the important role of the explicit knowledge's quality in trainable end-to-end dialogue systems. 
To support this investigation, we propose the Conversational-Logic-Injection-in-Neural-Network system (CLINN). CLINN stems from Domain Aware Multi-Decoder network \cite{Zhang2020}, which is a state-of-the-art trainable end-to-end task-oriented dialogue system. To allow explicit control of its dialogue state tracker and dialogue policy manager, CLINN includes a dedicated symbolic semi-logic language, in line with \citet{Jhunjhunwala2020}.
\completare{We also use CLINN in order to investigate the quality of the knowledge produced by a team of differently-skilled conversational designers.} We experimented with the \emph{Restaurant topic} of the MultiWOZ dataset \cite{Budzianowski2020MultiWozModelling}. We used two different sets of dialogues to allow conversational designers to generate explicit rules. Results show that rules injected are effective in the situation when training data are scarce and, moreover, experience in writing conversational rules is extremely important. In fact, rules from expert conversational designers used in CLINN significantly outperform a state-of-the-art neural-based dialogue system.

\section{Background and Related Work}
%Dialog systems scenario is mainly composed of two classes: task-oriented dialog systems and chatbots.
%The substantial difference that divides the two classes concerns into the interaction human-machine and their final scope.

Task-oriented dialogue systems are gaining an impressive attention in several real scenarios. However, when dialog systems are evaluated in settings with real users \cite{Laranjo}, their underlying models show up all their limitations. 

A specific study has shown the limitations of traditional rule-based dialogue systems in the health domain 
%of mental and physical health
\cite{Miner2016} when evaluated by external research groups. Devising strategies to generate more effective dialogue systems is then a clear need.

%The impressive study of \cite{Laranjo}  focuses on dialog systems in the medical domain and takes into consideration systems which have been evaluated in settings with real users. All analyzed systems are described as to have excellent results according to the analysis of the authors, which, presumably, are the producers of the systems. Instead, if the analysis is conducted by a research team which is independent with respect to the producers of the systems, conversational agents are unsatisfactory when used in the domain of mental and physical health \cite{Miner2016}.

%The aim of task-oriented dialog systems is to use user's conversations to help complete tasks, to obtain actions (such as booking request), to bring customer service or for workplace assistants.
%In the last decade, this aspect has increasingly fascinated industries seeking to create and improve Conversational AI task-oriented systems.
%All modern task-oriented dialogue systems are based around \textit{frames} and \textit{slots}.
%A frame is a kind of knowledge structure: it consists in a collection of slots each of which can take a set of possible values.
%Therefore a frame represents the kids of intentions that the system can extract from user's sentences. On the other hand, the set of slots in a frame specifies what the system needs to know and the filler of each slot is constrained to values of a particular semantic type. 

Learning end-to-end dialogue systems seems to be the path to go but huge annotated training sets are needed. Moreover, it is difficult to build up datasets in order to cover the expected distribution of dialogues in the target domain. It turns out that these datasets are quite sparse \cite{Budzianowski2020MultiWozModelling,kim2017fourth}. Alternative ways to help training neural-based dialogue systems are then gaining attention.

%For these reasons researchers began to thinks about adding knowledge or new methods for human-machine interactions thus introducing the concept of knowledge-based dialog systems.

Reinforcement learning is often used to reduce the centrality of annotated datasets.
%in learning neural-based dialogue systems. 
A fairly interesting approach is using %One of the most used choices is to train dialog systems using reinforcement learning algorithms through 
an Agenda-Based User Simulator (ABUS) \cite{Liu2017,Schatzmann2007Agenda-basedSystem}, which avoids introducing real humans in the learning loop. The advantage of using a user simulator is to get good performance without collecting data for supervised dialogue policy -- an expensive and time-consuming process. The basic idea of an ABUS model is to build hand-crafted rules according to an agenda which is declared before the dialogue started.
ABUS has been used in different domains such as  %we recall for example the hidden agenda user simulator \cite{schatzmann2007agenda}, the application of ABUS in
the movie domain \cite{li2016user}.
%and the implementation of six independent user simulator trained with different dialog planning and generation methods \cite{shi2019build}.
Nevertheless, there is no standard automatic metric for evaluating these user simulators, as it is unclear to define how closely the simulator resembles real user behaviors.
Indeed, although there are standards metrics to evaluate a user simulator under different aspects \cite{kobsa1994user,chin2001empirical}, there is no metric that actually correlates performance of a user simulator with human satisfaction \cite{shi2019build}.

%The annual DST-challenge called Dialog System Technology Challenge (DSTC) and the growing use of neural networks have changed the typical structures used for DST models.
%Conversational neural network, for example, has been used to learn utterance representation \cite{wen2016network} or a pointer network with a Seq2Seq architecture were generated to handle unseen slot values \cite{xu2018end}.
%A pre-trained BERT model \cite{devlin2018bert} has been used to encode slots and utterances and  also to find relevant information in dialogue context for predicting slot values using its multi-head attention \cite{vaswani2017attention} \cite{lee2019sumbt}.

A more direct way to introduce knowledge in neural-based dialogue systems is by injecting rules in dialogue state trackers and in dialogue policy managers. This approach is the more general line of research of merging symbolic knowledge and neural networks. In the context of neural-based dialogue systems, this line is pursued by using constrained rules 
%Most recently, researchers move towards interactive learning models with human-machine interactions 
\cite{Jhunjhunwala2020}, logical rules to be used in inductive logic programming  \cite{Zhou2020ResourceProgramming} or declarative language \cite{Altszyler2020}.
These rules and models can be easily included in the existing dialogue state tracking models %(such as neural network architectures) 
to guide the training and predictions phases without additional learning parameters \cite{hu2016harnessing,van2020analyzing}.
These models obtain the same advantage of the user simulator and in addition overcome the problem of the evaluation of the user-simulator itself.
Indeed the injected knowledge is, in different ways, rules governed by conversational designers.

\completare{However, there is not an extensive study on how these additional rules introduced by conversational designers affect the performance of the overall system.} %However, there is not an extensive study on how the quality of these additional rules introduced by conversational designers affect the performance of the overall system.

\begin{figure*}
    \centering
    \includegraphics[height=8.0cm]{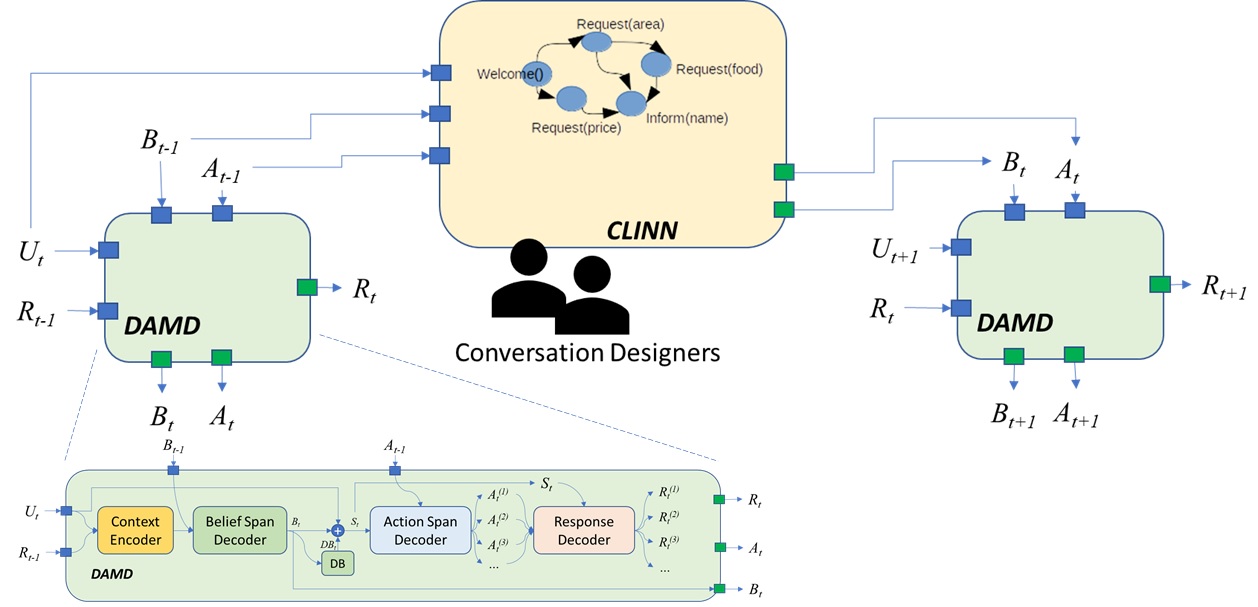}%/OUR_SOLUTION.jpg}
    \caption{Injecting External Knowledge with \OURSYS{} and Architecture of the Domain Aware Multi-Decoder (DAMD) network}
    \label{fig:OurSolution}
\end{figure*}

\section{Method and System} 

This section introduces our model to inject knowledge of conversational designers in end-to-end dialogue systems. Our solution is the Conversational-Logic-Injection-in-Neural-Network system, which is a knowledge-based dialogue state tracker, and it is used in combination with the Domain Aware Multi-Decoder network (DAMD), which is a state of the art end-to-end dialogue systems. We here describe first DAMD and then CLINN. 

\subsection{Domain Aware Multi-Decoder
(DAMD) network}
\writers{Fabio}{Giancarlo, Cristina} 

The Domain Aware Multi-Decoder network (DAMD) \cite{Zhang2020} offers a great opportunity to inject external knowledge. 

In fact, DAMD produces symbolic representations of dialogue states at each turn of the dialogue. 
Dialogue states $S_t$ at the time $t$ are triples $(R_t,B_t,A_t)$ where $B_t$ is the belief state, $A_t$ is the selected action and $R_t$ is the answer of the system given the action $A_t$. 
The symbolic representation of these states is based on  
%In this study, we use an end-to-end dialog architecture that includes the concept of 
\emph{belief spans} \cite{Lei2018}. These belief spans are sequences of symbols expressing belief states, which are the inner parts of dialogue states.
%at each turn of the dialogue. 

So DAMD is a module-based neural network that, at a given time $t$, takes as input $S_{t-1}$ and $U_t$ and produces $S_t$ where $U_t$ is the analyzed user utterance. DAMD consists of four seq-to-seq modules plus the access to an external database (see Fig. \ref{fig:OurSolution}). 
%The pipeline is applied for each turn of the dialogue. It, globally, takes four inputs $(U_t,R_{t-1},B_{t-1},A_{t-1})$ and produces three outputs $(R_{t},B_{t},A_{t})$ where $t$ is the actual turn, $U_t$ is the user utterance,  $R_{t-1}$ and $R_{t}$ are the previous and the current system responses, $B_{t-1}$ and $B_{t}$ are the previous and the current belief state spans, $A_{t-1}$ and $A_{t}$  are the previous and the produced system actions. 
The four modules behave as follows: The \emph{context encoder} encodes the context of the turn $(U_t, R_{t-1})$ in a context vector $c_t$. The \emph{belief span decoder} decodes the previous belief span $B_{t-1}$ and, along with the context vector $c_t$ produces the belief span $B_t$ of current turn. This $B_t$ is used to query the database $DB$ and the answer $DB_t$ is concatenated with $B_t$ to form the internal state $S_t$ of the turn. Then, the \emph{action span decoder} produces the current action $A_t^{(i)}$ by taking into consideration the current state $S_t$ and the previous action $A_{t-1}$ . Finally, the \emph{response decoder} emits the final response $R_t^{i}$ taking into consideration the current state $S_t$ and the corresponding action $A^{(i)}_{t}$. In \cite{Zhang2020}, multiple actions and multiple responses are produced to increase variability in dialogues and, for this reason, the framework is called multi-action data augmentation.

%\meta{Note:
In this work, we decided to use a simplified version of the DAMD architecture, which receives in input the user's dialogue act $U_t$ and the system action $A_{t-1}$ instead of the system response $R_{t-1}$. Moreover, we removed the response decoder, leaving a simple action decoder that generates a single action $A_t$.

%\begin{figure*}
%    \centering
%
%    \includegraphics[width=13cm]{imgs/MADAorg.jpg}
%    \caption{Architecture of the Domain Aware Multi-Decoder
%(DAMD) network}
%    \label{fig:MADAorg}
%\end{figure*}

%\subsection{Short description of the proprietary system}
%\writers{Cristina}{Fabio, Giancarlo} 

\subsection{Hand-crafting a Dialogue State Tracker}
\writers{Cristina}{Fabio, Giancarlo} 

%DAMD network offers a tremendous opportunity to inject external knowledge. In fact, the \emph{belief span decoder} transforms the internal context vector $c_t$ and an explicit symbolic previous belief span $B_{t-1}$ in an explicit belief span $B_{t}$. In the same way, the action span decoder takes in input an explicit, symbolic previous action $A_{t-1}$. As $B_{t-1}$ and $A_{t-1}$ are explicit, these can be easily controlled by an external, symbolic module.    

%\subsubsection{Hand-crafted Dialog State Tracker}
Within DAMD, we then propose a knowledge based dialogue state tracker, that is, our Conversational Logical Injection in Neural Network (\OURSYS{}). 

\OURSYS{} allows conversational designers to control the dialogue flow with symbolic transition rules. It is a fully operational dialogue state tracker, which may evolve by itself (CLINN base) or may work in cooperation with DAMD (CLINN+DAMD) by substituting dialogue states when its transition rules fire (see Fig. \ref{fig:OurSolution}).

In the following, we describe the representation of dialogue states and of transition rules in a semi-logical language.

\subsubsection{Dialogue States in a Semi-logical Language}

As far as we aim to inject knowledge in the dialogue system, we need to clarify how dialogue states are represented and how rules can be described. For this reasons, we express both dialogue states and rules in a logical form (as in \cite{Zhou2020ResourceProgramming}). By using this logical form, rules can be expressed by using logical constraints and variables.

The shared representation of dialogue states (c.f., \cite{Henderson2020}) is then presented in the following way:
\begin{center}
$S_t = $ 
\begin{tabular}{ll}
 \hline
    $U_t$ & Inform(food(thai)) \\
               & Inform(time(15:00)) \\
               & Request(name) \\
 \hline
    $B_{t-1}$ & area(west) \\
 \hline
    $A_{t-1}$ & Request(food) \\
 \hline
\end{tabular}
\end{center}
In belief states $B_{t-1}$, logical facts are represented as feature(value), for example, area(west). Instead, in the case of user utterances $U_t$ and system actions $A_{t-1}$, feature-value pairs are inserted in predicates representing dialogue acts and, then, are represented as dialogue\_act(feature(value)), for example, Inform(time(15:00)). This representation is needed to indicate dialogue acts hidden in user utterances and in system actions. The catalogue of dialogue acts we are using is presented in the Experimental section (Table \ref{tab:dialog_acts}).

\subsubsection{Transition Rules in a Semi-logical Language}

Transition rules for controlling the dialogue state tracker are then expressed in a logic programming formalism, that is, horn clauses with variables. For the sake of simplicity, these transitions rules are expressed as \emph{preconditions} and \emph{action}. 
Preconditions are matched on the current dialogue states. If preconditions fire and variables are unified with current values, the result of the rule is to add or replace the bounded action in the next dialogue state. In the following, there are two example transition rules:
\begin{center}
\begin{tabular}{p{3.5cm}lp{3cm}}
  \emph{Preconditions}  & \hspace{-3mm} $\rightarrow$ & \hspace{-3mm} \emph{Action} \\
\hline
\hline
$R_1 = $ \hspace{-5mm} \\
\begin{tabular}{ll}
    $U_t$ & Inform(food(X)) \\
 \hline
    $B_{t-1}$ & area(Y) \\
 \hline
    $A_{t-1}$ & Request(food) \\
\end{tabular} &
 \hspace{-3mm} $\rightarrow$   & 
 \hspace{-3mm} \begin{tabular}{ll}
 \hline
    $B_{t}$ & area(Y) \\
            & food(X) \\
 \hline 
\end{tabular} \\
\hline
\hline
\end{tabular}
\end{center}
\begin{center}
\begin{tabular}{p{3.5cm}lp{3cm}}
  \emph{Preconditions}  & \hspace{-3mm} $\rightarrow$ & \hspace{-3mm} \emph{Action} \\
 \hline 
 \hline 
 $R_2 = $\\
 \begin{tabular}{ll}
 $U_t$ & Inform(food(X)) \\
    & Request(address) \\
 \hline
    $B_t$ & area(Y), food(X) \\
 \hline
    $A_{t-1}$ & Request(food) \\
 \hline
    $DB_t$ & between(4,10)\\
\end{tabular} &\hspace{-3mm}
$\rightarrow$ & 
 \hspace{-3mm} \begin{tabular}{ll}
\hline    
    $A_{t}$ & Inform(address)\\
        & Request(price) \\
\hline
\end{tabular}\\
 \hline 
 \hline
\end{tabular}
\end{center}
Given the above state $S_t$, the transition rule $R_1$ fires. In fact, all its preconditions are satisfied and the variables $X$ and $Y$ are unified to the values $thai$ and $west$ as $Inform(food(X))$ in $U_{t}$ is matched to $Inform(food(thai))$ and $area(Y)$ in $B_{t}$ is matched to $area(west)$. As result of the application of this transition rule, $area(west)$ and $food(thai)$ are added to the belief or override existing believes. 
The application of the transition rule $R_2$ is similar but its effect is on the next action of the system $A_t$. 

During transition rule production, we asked conversational designers to build up two types of rules: rules affecting only the belief $B_t$ (belief rules) and rules affecting only the action $A_t$ (action rules). Belief and action rules are then sorted in two separated lists and the selection algorithm takes the first rule for each type whose constraints are satisfied.

Transition rules, as written by conversational designers, may be over-constrained and this fact may hinder its application in novel dialogues. For this reason,  we foresee two ways of application of transition rules:
\begin{itemize}
    \item Fully constrained (\emph{Full}) - all constraints are considered 
    %\item Partially constrained (Part) - the constraint on previous action is not considered in belief rules
    \item Partially constrained (\emph{Free}) - constraints on previous actions are not considered in belief rules and  constraints on current beliefs are not considered in action rules 
\end{itemize}
In the experimental section, we will analyze how this may affect the final performance of the dialogue system.

%    User(Inform;food=X), PrevBelief(area=Y), PrevSystem(Request;food)  $\rightarrow$  Belief(area=Y,food=X) 

%\completare{Andrebbe inserito qualcosa qui visto tra la section e la subsection? ad esempio il perchè si chiamano PRELIMINARY exp?}

\begin{table}[h]
%\hline
%\hline
    \centering
    \begin{tabular}{p{6.5cm}}
    \textbf{General-domain}\\
         bye, greet, reqmore , welcome 
    \end{tabular}
\vspace{5mm}
    \begin{tabular}{p{6.5cm}}
    \textbf{Restaurant-domain}\\
         inform, request, nooffer, recommend, select, offerbook, offerbooked, nobook
    \end{tabular}
%\hline
%\hline
\\
    \textbf{Additional User Dialogue Acts}\\
    \begin{tabular}{cp{4cm}}
         \emph{Act} &  \emph{Description} \\
         \hline
 getrecommend & asking for a recommendation\\
 acceptance & accepting system's proposals\\
 rejection & reject system's proposals\\
 alternatives & asking for other restaurants\\
    \end{tabular}
%\hline
%\hline
    \caption{General and Domain Specific Dialogue Acts from MultiWoZ, along with our additional dialogue acts.}
    \label{tab:dialog_acts}
\end{table}

\section{Experiments}
%mic: Cambierei la seguente frase in: 
Through these experiments, our aim is to answer the following questions:
%In these experiments, we aim to answer to our research questions: 
\completare{(1) how important is external knowledge in neural-based dialogue systems and (2) how relevant is the difference in skills of conversational designers involved in the production of transition rules.} %(1) how important is external knowledge in neural-based dialogue systems and (2) how relevant is the expertise of conversational designers involved in the production of transition rules. 
Hence, this experimental section is organized as follows: Section \ref{sec:set_up}
% mic: cambierei il 'describes' in shows poichè dopo 6 parole viene ripetuto il medesimo verbo
shows the setting of our experiments by describing the general principles, the dialogue corpus, the production of transition rules, the evaluation of the coherence of transition rules produced by different conversational designers and the metrics used to evaluate the dialogue systems. Section \ref{sec:discussion} analyses and discusses results.

\begin{table*}[t]
\centering
\resizebox{14cm}{!}{%{7.8cm}{!}{%
\renewcommand{\arraystretch}{1.00}%{1.00}
\begin{tabular}{|P{1.6cm}|P{1.6cm}|c|c|c|c|}
\hline
\multicolumn{2}{|c|}{}                              & \multicolumn{4}{c|}{\textbf{Inter-designer agreement}}                         \\ \cline{3-6} 
\multicolumn{2}{|c|}{}                              & \multicolumn{2}{c|}{\textbf{Small set}} & \multicolumn{2}{c|}{\textbf{Medium set}} \\ \cline{3-6} 
\multicolumn{2}{|c|}{\multirow{-3}{*}{\textbf{\makecell{Pair of conversational \\designers}}}}  &\textbf{Belief rules}       &\textbf{Action rules}           &\textbf{Belief rules}           &\textbf{Action rules}                    \\ \hline
% %\cellcolor{expert}expert & \cellcolor{medium}junior1 &    X       &      X     &     X      &    X       \\ \hline
% \cellcolor{expert}expert & \cellcolor{low}junior1 &    0.51       &      0.26     &  0.52         &      0.30     \\ \hline
% \cellcolor{expert}expert & \cellcolor{low}junior2 &     0.68      &      0.28     &  0.58         &      0.30     \\ \hline
% %\cellcolor{medium}A2 & \cellcolor{low}A3 &      X     &     X      &    X       &       X    \\ \hline
% %\cellcolor{medium}A2 & \cellcolor{low}A3 &      X     &     X      &    X       &       X    \\ \hline
% \cellcolor{low}junior1 & \cellcolor{low}junior2 &     0.77      &     0.71      &     0.88      &     0.74      \\ \hline
\cellcolor{expert}exp1 & \cellcolor{expert}exp2 & 0.52 & 0.36 & 0.49 & 0.43\\ \hline
\cellcolor{expert}exp1 & \cellcolor{low}jr1 & 0.31 & 0.17 & 0.55 & 0.23\\ \hline
\cellcolor{expert}exp1 & \cellcolor{low}jr2 & 0.39 & 0.26 & 0.48 & 0.33\\ \hline
\cellcolor{expert}exp1 & \cellcolor{low}jr3 & 0.53 & 0.41 & 0.53 & 0.39\\ \hline
\cellcolor{expert}exp2 & \cellcolor{low}jr1 & 0.38 & 0.14 & 0.64 & 0.18\\ \hline
\cellcolor{expert}exp2 & \cellcolor{low}jr2 & 0.57 & 0.26 & 0.54 & 0.29\\ \hline
\cellcolor{expert}exp2 & \cellcolor{low}jr3 & 0.76 & 0.29 & 0.61 & 0.3\\ \hline
\cellcolor{low}jr1 & \cellcolor{low}jr2 & 0.44 & 0.3 & 0.84 & 0.4\\ \hline
\cellcolor{low}jr1 & \cellcolor{low}jr3 & 0.53 & 0.37 & 0.97 & 0.53\\ \hline
\cellcolor{low}jr2 & \cellcolor{low}jr3 & 0.77 & 0.66 & 0.86 & 0.74\\ \hline
\end{tabular}
}
\begin{center}

\fbox{\begin{tabular}{cc}
Level of expertise: \textcolor{expert}{$\blacksquare$} High &
%\textcolor{medium}{$\blacksquare$} Medium expertise level& 
\textcolor{low}{$\blacksquare$} Low  
\end{tabular}
}

\end{center}
\caption {Inter-designer agreement score between pairs of annotators calculated on the rules constructed from small and medium set. The color of annotator entry mean his level of expertise}
\label{tab:Inter-annotator}
\end{table*}

\subsection{Experimental Set-Up}
\label{sec:set_up}

\subsubsection{General principles and Dialogue corpus}

We adopted a general principle in experiments involving learning and neural networks: performing repeated experiments and evaluating statistical significance of the difference among different configurations. Results of neural-based dialogue systems, as well as results of all experiments using neural networks, may vary a lot depending on the initial conditions. Different seeds given to random pseudo-generators can determine different initial conditions of the learning. 
%mic: cambiato "hence" in "therefore"
Therefore, we repeated each experiment involving DAMD for 6 times with 6 different fixed seeds. Whenever relevant, we report mean and standard deviations of results and we compute a paired statistical significance analysis. 

We evaluated the production of transition rules used by \OURSYS{} on a version of the MultiWOZ dataset \cite{Budzianowski2020MultiWozModelling} extended by \cite{lee2019convlab}, as in Zhang et al. \cite{Zhang2020}. This dataset is widely used and it has been designed as a human-human task-oriented dialogue dataset collected via the Wizard-of-Oz framework. 
%mic: questa frase non mi convince ma non so come cambiarla
One participant plays the role of the system. The dataset contains conversations on several domains in the area of touristic information (hotel, train, restaurant, taxi,...).
%mic: cambierei la seguente frase in: Each domain has a set of dialog acts and some greeting phrases like "goodbye" or "thanks"
Each domain has a set of dialogue acts in addition to some general acts such as \emph{greeting} or \emph{bye}. 
Users' and system's interactions are described in term of these dialogue acts. 
%We also annotated missing user's dialogue acts in the dataset. For some cases, we introduced some additional dialogue acts, which were, in our opinion, more suitable. Additional dialogue acts are listed and described in Table~\ref{tab:dialog_acts}.

We focused on the restaurant domain of the MultiWOZ dataset that consists of 1200 dialogues for the training set, 61 dialogues for the testing set and 50 dialogues for the validation set. 
To simulate data scarcity at different levels, we derived three additional training sets by randomly sampling the full training sets. These additional training sets contain 150, 300 and 450 dialogues. So, results will be presented both on specific training set and by using a sort of training curve with respect to the increasing number of dialogues. 

Finally, we annotated missing user's dialogue acts in the dataset. For some cases, we also introduced some additional dialogue acts, which are, in our opinion, more suitable. Additional dialogue acts are listed and described in Table~\ref{tab:dialog_acts}.

%Finally, we modified the MultiWoz dataset \cite{Budzianowski2020MultiWozModelling} by annotating missing user's dialogue acts. The final annotations are similar to other dialogue acts used in the dataset; in some cases they have additional dialogue acts we defined, which were, in our opinion, more suitable for some utterances. Additional dialogue acts are listed and described in Table~\ref{tab:dialog_acts}.
%The final annotations are similar to other dialogue acts used in the dataset; in some cases they have additional dialogue acts we defined, which were, in our opinion, more suitable for some utterances. Additional dialogue acts are listed and described in Table~\ref{tab:dialog_acts}.
%Finally, we modified the MultiWOZ dataset \cite{Budzianowski2020MultiWozModelling} for taking into consideration additional dialogue acts, which, in our opinion, were missing although important for better modeling dialogs. These are listed and described in Table~\ref{tab:dialog_acts}

 %Note: The transition rule does not fire since the $R_1$ does not exactly cover $S_t$. Modify the $S_t$ example. The Horn clause is not appropriate, since we perform an exact match \dots
 In our experiments, CLINN only applies rules whose preconditions exactly match with the current dialogue state (e.g. state's user act must be exactly as the precondition's user act).

%QUESTI SONO QUELLI CHE AVETE AGGIUNTO? [alternatives, getrecommend, acceptance, rejection]: these are DActs added and used only for user sentences whose dacts were empty.
\subsubsection{Editing and Evaluating Transition rules in Dialogue State Trackers}
%mic: Seppur "expertise" è la forma corretta per indicare nel nostro caso l'esperienza di un utente, nell'elenco che segue la forma giusta è "experience"

Our goal is twofold: (1) understanding whether external explicit knowledge is useful in neural-based dialogue systems in poor settings and (2) determining the effect of the skills of conversational designers in the performance of neural-based dialogue systems.
%determining the effect of the experience of conversational designers in the performance of neural-based dialogue systems.

\completare{
Hence, we built up a team of five conversational designers -- two experts, and three juniors -- with different level of expertise:
 \begin{itemize}
     \item the experts (exp1, exp2) have more than 15 years of experience in natural language processing and more than 5 years of experience in experimental and production dialogue systems   
     \item the juniors (jun1, jun2, jun3) have less than 1 year of experience in NLP and no experience in dialogue systems production
 \end{itemize}
The three low-expertise-level conversational designers have been trained for a week before producing the transition rules.
}
%--------------------------------------------------
% Hence, we built up a team of three conversational designers -- expert, junior1 e junior2 -- with different level of expertise:
%  \begin{itemize}
%      \item expert has more than 15 years of experience in natural language processing and more than 5 years of experience in experimental and production dialogue systems   
%      \item junior1 and junior2 have less than 1 year of experience in NLP and no experience in dialogue systems
%  \end{itemize}
% The two low-expertise-level conversational designers, junior1 and junior2, have been trained for a week before producing the transition rules.

%mic: La frase è troppo lunga: la dividerei in punti del tipo: 

The procedure to build the sets of transition rules is the following. Each conversational designer generated a transition rules sets using the following two steps: 1) s/he observes a set of training examples extracted from the corpus; 2) s/he generates a set of transition rules in order to model the expected conversational behavior of the user. A direct consequence of 2) will be the modeling of the dialogue system's responses.
%each conversational designer observes a set of training examples extracted from the corpus and, by using its own intuition on the domain, generates a set of transition rules in order to model the expected conversational behavior of the user and, consequently, of the answers of the dialogue system.
We asked conversational designers to produce two separate sets of transition rules: $bs\_rules$ for the production of the belief state $B_t$ and $action\_rules$ for the production of the system action $A_t$.  
Conversational designers are exposed to two sets of dialogues derived from the training dataset: \emph{small set} and the \emph{medium set}. The \emph{small set} contains 5 dialogue examples while the \emph{medium set} contains the small set adding another 10 conversation examples. Clearly, designers see the small set and produce a first set of transition rules and, only then, they see the \emph{medium set} to produce the second set of transition rules. At the end, we can experiment with 6 different transition rule sets and each rule set has two sections, one for the belief and one for the action. \meta{TODO: specify how much time annotators spent in rule construction}

After the rule-sets construction, we measured the inter-designer agreement for each pair of annotators in order to validate the design rules policies , one measure for $bs\_rules$ and another one for $action\_rules$. We used an inter-designer agreement measure:
\begin{equation}
    AGR = \frac{|R1 \cap R2|}{|R1 \cup R2|} 
\end{equation}
where $R1$ and $R2$ are the sets of rules produced by the first and second annotator respectively.

%SAMIRRRRRRR dimmi quando posso operare :-D :-D hai letto????????? Stai dormendo su un video youtube? Credo di si :-D 

%The generated rule-sets were injected into DAMD network using CLINN approach. To analyze the difference between the results based on conversational designers experience, we evaluate the model performances for each conversational designer rule-sets.

%%%%%%%%%%%%%%%%%%%%%%%%%%%%%%%%%%%%%%%%%%%%%%%%%%%%%%
%New table containing comparison between Full (Constrained) and Free Rule types
%this table shows a bigger gap between restricted and flexible rules
\begin{table*}[t]%[htbp]
\centering
\resizebox{14cm}{!}{%{12cm}{!}{%{7.8cm}{!}{%
\renewcommand{\arraystretch}{0.7}%{1.00}
\begin{tabular}{cccccccc}
\hline
%\multirow{2}{*}{\textbf{Model}} & \multicolumn{3}{c}{\textbf{Rule Set}} &  \multicolumn{4}{c}{\textbf{Metrics}}  \\  %\multirow{2}{*}{\textbf{Action F1}} & \multirow{2}{*}{\textbf{Joint Goal}} \\
\multicolumn{1}{c|}{\multirow{2}{*}{\textbf{Model}}} & \multicolumn{3}{c|}{\textbf{Rule Set}} &  \multicolumn{4}{c}{\textbf{Metrics}}  \\  %\multirow{2}{*}{\textbf{Action F1}} & \multirow{2}{*}{\textbf{Joint Goal}} \\
\multicolumn{1}{c|}{} & \textbf{Size} & \textbf{Type} & \multicolumn{1}{c|}{\textbf{Designer}} & \textbf{Action F1} & \textbf{Joint Goal} & \textbf{Slot Acc} & \textbf{Slot F1}\\ \hline \hline
\textbf{Fully-informed DAMD} &  &  &  & 44.8  (±2.02) & 72.23 (±1.49) & 98.37 (±0.17) & 92.92 (±0.67) \\
\textbf{DAMD} & \textbf{} &  &  & 43.75  (±1.59) & 56.67 (±4.34) & 97.17 (±0.48) & 87.6 (±2.08) \\ \hline
\multicolumn{1}{c|}{\multirow{20}{*}{\textbf{CLINN base}}} & \multirow{10}{*}{\textbf{small}} & \multirow{5}{*}{Full} & exp1 & 14.6 & 28.1 & 92.7 & 57 \\
\multicolumn{1}{c|}{} &  &  & exp2 & 13.6 & 30.9 & 92.6 & 56.4 \\
\multicolumn{1}{c|}{} &  &  & jr1 & 8.9 & 17.3 & 90.9 & 39.3 \\
\multicolumn{1}{c|}{} &  &  & jr2 & 8.9 & 25.5 & 92.5 & 54.9 \\
\multicolumn{1}{c|}{} &  &  & jr3 & 8.9 & 25.5 & 92.5 & 54.9 \\
\multicolumn{1}{c|}{} &  & \multirow{5}{*}{Free} & exp1 & 14.7 & 38.8 & 93.8 & 66.9 \\
\multicolumn{1}{c|}{} &  &  & exp2 & 14.5 & 38.8 & 93.8 & 66.9 \\
\multicolumn{1}{c|}{} &  &  & jr1 & 12.9 & 24.8 & 91.9 & 50.5 \\
\multicolumn{1}{c|}{} &  &  & jr2 & 14.9 & 38.8 & 93.8 & 66.9 \\
\multicolumn{1}{c|}{} &  &  & jr3 & 14.7 & 38.8 & 93.8 & 66.9 \\ \cline{2-8} 
\multicolumn{1}{c|}{} & \multirow{10}{*}{\textbf{medium}} & \multirow{5}{*}{Full} & exp1 & 19.4 & 29.9 & 93 & 59.8 \\
\multicolumn{1}{c|}{} &  &  & exp2 & 18.2 & 27.3 & 92.7 & 57.7 \\
\multicolumn{1}{c|}{} &  &  & jr1 & 8.9 & 25.2 & 92.4 & 56.2 \\
\multicolumn{1}{c|}{} &  &  & jr2 & 8.9 & 25.2 & 92.4 & 56.2 \\
\multicolumn{1}{c|}{} &  &  & jr3 & 8.9 & 25.2 & 92.4 & 56.2 \\
\multicolumn{1}{c|}{} &  & \multirow{5}{*}{Free} & exp1 & 20.1 & 44.6 & 94.8 & 72.3 \\
\multicolumn{1}{c|}{} &  &  & exp2 & 19.1 & 45.3 & 94.9 & 73 \\
\multicolumn{1}{c|}{} &  &  & jr1 & 20 & 45.7 & 95.2 & 75.1 \\
\multicolumn{1}{c|}{} &  &  & jr2 & 19.2 & 45.7 & 95.2 & 75.1 \\
\multicolumn{1}{c|}{} &  &  & jr3 & 20 & 45.7 & 95.2 & 75.1 \\ \hline \hline
\multicolumn{1}{c|}{\multirow{20}{*}{\textbf{CLINN}}} & \multirow{10}{*}{\textbf{small}} & \multirow{5}{*}{Full} & exp1 & 43.38  (±1.56) & 57.6 (±4.87)$\dagger$ & 97.23 (±0.5)$\diamond$ & 87.9 (±2.12)$\dagger$ \\
\multicolumn{1}{c|}{} &  &  & exp2 & 43.28  (±1.42) & 57.97 (±5.3)$\diamond$ & 97.25 (±0.54)$\diamond$ & 88 (±2.26)$\diamond$ \\
\multicolumn{1}{c|}{} &  &  & jr1 & 43.78  (±1.7) & 57.25 (±4.5)$\dagger$ & 97.23 (±0.49)$\diamond$ & 87.87 (±2.11)$\star$ \\
\multicolumn{1}{c|}{} &  &  & jr2 & 43.7  (±1.57) & 57.25 (±4.5)$\dagger$ & 97.23 (±0.49)$\diamond$ & 87.87 (±2.11)$\star$ \\
\multicolumn{1}{c|}{} &  &  & jr3 & 43.77  (±1.69) & 57.25 (±4.5)$\dagger$ & 97.23 (±0.49)$\diamond$ & 87.87 (±2.11)$\star$ \\
\multicolumn{1}{c|}{} &  & \multirow{5}{*}{Free} & exp1 & 43.38  (±1.66) & 61.5 (±3.82)$\star$ & \textbf{97.48 (±0.41)$\star$} & \textbf{88.92 (±1.77)$\star$} \\
\multicolumn{1}{c|}{} &  &  & exp2 & 43.38  (±1.56) & 61.43 (±4.25)$\star$ & 97.47 (±0.43)$\star$ & 88.85 (±1.97)$\star$ \\
\multicolumn{1}{c|}{} &  &  & jr1 & \textbf{44.28  (±2.04)} & \textbf{61.52 (±3.78)$\star$} & 97.47 (±0.43)$\star$ & 88.87 (±1.94)$\star$ \\
\multicolumn{1}{c|}{} &  &  & jr2 & 43.12  (±1.89) & 61.18 (±4)$\star$ & 97.43 (±0.46)$\diamond$ & 88.77 (±1.96)$\star$ \\
\multicolumn{1}{c|}{} &  &  & jr3 & 43.18  (±1.78) & 61.18 (±4)$\star$ & 97.43 (±0.46)$\diamond$ & 88.77 (±1.95)$\star$ \\ \cline{2-8} 
\multicolumn{1}{c|}{} & \multirow{10}{*}{\textbf{medium}} & \multirow{5}{*}{Full} & exp1 & \textbf{45.07  (±2.16)$\star$} & 59.6 (±4.45)$\star$ & 97.32 (±0.47)$\star$ & 88.23 (±2.16)$\star$ \\
\multicolumn{1}{c|}{} &  &  & exp2 & 44.8  (±2.02)$\star$ & 59.55 (±2.84)$\diamond$ & 97.33 (±0.46)$\diamond$ & 88.33 (±2.01)$\star$ \\
\multicolumn{1}{c|}{} &  &  & jr1 & 43.52  (±1.92) & 57.08 (±3.9) & 97.2 (±0.4) & 87.82 (±1.77) \\
\multicolumn{1}{c|}{} &  &  & jr2 & 43.45  (±1.49) & 57.08 (±3.9) & 97.2 (±0.4) & 87.82 (±1.77) \\
\multicolumn{1}{c|}{} &  &  & jr3 & 43.53  (±1.97) & 57.08 (±3.9) & 97.2 (±0.4) & 87.82 (±1.77) \\
\multicolumn{1}{c|}{} &  & \multirow{5}{*}{Free} & exp1 & 44.95  (±2.51)$\dagger$ & \textbf{64.13 (±1.98)$\star$} & \textbf{97.68 (±0.24)$\star$} & \textbf{89.83 (±1)$\star$} \\
\multicolumn{1}{c|}{} &  &  & exp2 & 44.73  (±2.36)$\dagger$ & 62.5 (±1.95)$\dagger$ & 97.62 (±0.25)$\star$ & 89.63 (±1.16)$\star$ \\
\multicolumn{1}{c|}{} &  &  & jr1 & 44.75  (±2.69)$\dagger$ & 61.63 (±1.42)$\diamond$ & 97.55 (±0.26)$\dagger$ & 89.32 (±1.11)$\dagger$ \\
\multicolumn{1}{c|}{} &  &  & jr2 & 44.45  (±1.97)$\star$ & 61.63 (±1.42)$\diamond$ & 97.55 (±0.26)$\dagger$ & 89.3 (±1.08)$\dagger$ \\
\multicolumn{1}{c|}{} &  &  & jr3 & 44.72  (±2.66)$\dagger$ & 61.63 (±1.42)$\diamond$ & 97.55 (±0.26)$\dagger$ & 89.3 (±1.08)$\dagger$ \\ \hline
\end{tabular}%
}
\caption{Experimental results using a training set of 300 examples. Mean and standard deviation are shown for each model configuration. Results are obtained from 6 runs, and the symbols $\dagger$,$\diamond$,$\star$ indicate a statistically significant better score than DAMD with a confidence level of, respectively, 90\%,95\%,99\% with the sign test.}
\label{tab:pact-results}
\end{table*}

\subsubsection{Evaluation metrics for Dialogue systems}
The automatic evaluation of dialogue systems is, in general, a very difficult problem \cite{EvalDialogueMetrics}. Yet, since a human evaluation is extremely expensive, we used measures widely adopted to evaluate both dialogue state trackers and actions of dialogue systems. These measures, hereafter described, are: \emph{Action-F1}, \emph{Slot Accuracy}, \emph{Joint Goal} and \emph{Slot F1}.   

%We evaluated our model using the following metrics:
%[label={\tiny\raisebox{1ex}
%\renewcommand{\labelitemi}{$\bullet$}    
%\renewcommand{\labelitemii}{$\blacksquare$}
%\begin{itemize}
\noindent \textbf{Action-F1}: the micro-averaged F1-score of the predicted dialogue action $a_t$ compared to the correct one $\hat{a_t}$. %this metric checks if the predicted dialogue action $a_t$ matches the correct dialogue action $\hat{a_t}$.
    
\noindent \textbf{Slot Accuracy}: is defined as the fraction of slots values correctly predicted by the model over all slot values. For each dialogue turn $D_t$, the average slot accuracy is defined as follows:\\
    \begin{equation}%$$
        \frac{1}{n}\sum_{i=1}^{n} \mathbbm{1}_{y_i=\hat{y_i}}
    \end{equation}%$$
    where:
    %\begin{itemize}[label={*}]
    %    \item 
        $y_i$ is the predicted slot value, 
    %    \item 
        $\hat{y_i}$ is the correct slot value, 
        %\item 
        $n$ is the total number of slots and
        %\item 
        $\mathbbm{1}_{y_i=\hat{y_i}}$ is an indicator variable that is 1 if and only if $y_i = \hat{y_i}$.
    %\end{itemize}
    
\noindent \textbf{Joint-Goal}: the joint-goal is defined as the fraction of dialogue turns for which the values $v_i$ for all slots $s_i$ are predicted correctly. For each dialogue turn $D_t$, we compute:%The metric is defined by the following formula:
    \begin{equation}
         \mathbbm{1}_{(\sum_{i=1}^{n} \mathbbm{1}_{y_i=\hat{y_i}})=n}
    \end{equation}
    
\noindent \textbf{Slot F1}: is defined as the micro-averaged F1-score of slot prediction.
%\end{itemize}

\subsubsection{Configurations and Meta-parameters}

We experimented with four configurations: \emph{DAMD}, \emph{Fully-informed DAMD}, \emph{CLINN base}, \emph{CLINN} and \emph{CLINN-merge}.
\emph{DAMD} is the basic DAMD system tested in the configuration 
where DAMD evolves
where $B_{t-1}$ and $A_{t-1}$ at the step $t$ are the actual $B$ and $A$ produced by the previous application of DAMD.
\emph{Fully-informed DAMD}, also referred as DAMD oracle, is the basic DAMD system tested in the configuration where DAMD evolves where $B_{t-1}$ and $A_{t-1}$ at the step $t$ are taken from the ground truth. Hence, this represents the upper bound of our study. \emph{CLINN base} is a system that evolves only utilizing transition rules written by conversational designers. Finally, \emph{CLINN} is a combination of DAMD and the module that applies rules written by designers. %Finally, \emph{CLINN-merge} is a model using two different and merged rule sets.
%\meta{(i.e. from the ground truth)}

We trained DAMD mainly using the same hyperparameters used in \cite{Zhang2020}. Our version of DAMD has 3 encoders and 2 decoders based on single-layer bidirectional GRUs with hidden size of 100. Since our focus is only on the restaurant domain, DAMD relied on a vocabulary restricted on words of that domain. %Since we focused on the restaurant domain only, we used a vocabulary size of 1500. 
In order to understand how the model's performances change with a different number of training examples, we separately trained DAMD on different datasets of 150, 300, 450, 1200 training examples. Moreover, we performed 6 different runs for each configuration with 6 different fixed seeds. \completare{We trained each model for at least 30 epochs saving the best model's parameters according to performances on validation set.}
%[Machine Details]
%Each model has been trained on a single-gpu "on cloud" machine for 15-20 minutes.

%We maintained the same values for word embedding size, learning rate, 
%50 word embedding size, 0.005 learning rate, 128 batch size, 100 epochs with 30 epochs early stop, Adam optimizer and ; each encoder/decoder used for the encoding/generation of the sequences is based on a single-layer GRU with 100 hidden size. 

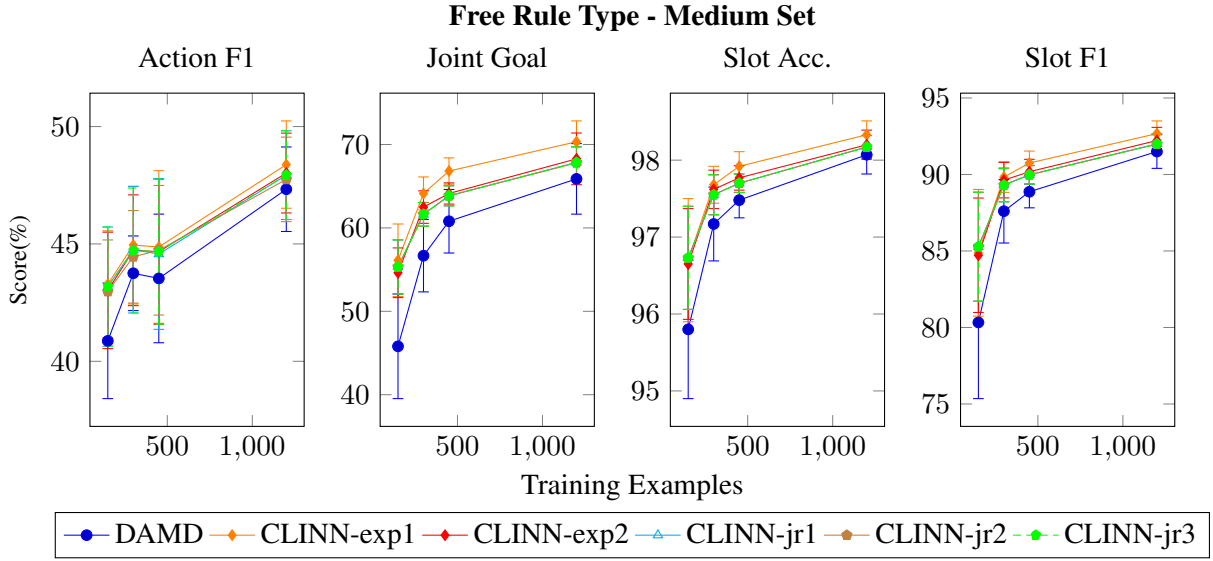
\begin{figure*}
    \centering
    \begin{tikzpicture}
     \begin{groupplot}[group style={group size= 4 by 1},height=6cm,width=4.4cm]%height=4.4cm,width=4.4cm]
    
    \nextgroupplot[title=Action F1,ylabel={\small Score(\%)},legend to name=models, legend columns=6]
    %medium action f1
    \addplot+[error bars/.cd, y dir=both,y explicit]
    coordinates {
    (150, 40.87) +- (0.0, 2.46)
(300, 43.75) +- (0.0, 1.59)
(450, 43.53) +- (0.0, 2.74)
(1200, 47.33) +- (0.0, 1.8)
    };\addlegendentry{DAMD}
    \addplot+[color=orange,mark=diamond*,mark options={orange},error bars/error bar style={orange},error bars/.cd, y dir=both,y explicit]
    coordinates {
    (150, 43.28) +- (0.0, 2.28)
(300, 44.95) +- (0.0, 2.51)
(450, 44.87) +- (0.0, 3.25)
(1200, 48.38) +- (0.0, 1.86)

    };\addlegendentry{CLINN-exp1}
    \addplot+[color=red,mark=diamond*,mark options={red},error bars/error bar style={red},error bars/.cd, y dir=both,y explicit]
    coordinates {
    (150, 43.02) +- (0.0, 2.48)
(300, 44.73) +- (0.0, 2.36)
(450, 44.67) +- (0.0, 3.09)
(1200, 48.02) +- (0.0, 1.7)

    };\addlegendentry{CLINN-exp2}
    % \addplot+[error bars/.cd, y dir=both,y explicit]
    % coordinates {};\addlegendentry{Intermediate}
    \addplot+[color=cyan,mark=triangle,mark options={cyan},error bars/error bar style={cyan},error bars/.cd, y dir=both,y explicit]
    coordinates {
    (150, 43.18) +- (0.0, 2.54)
(300, 44.75) +- (0.0, 2.69)
(450, 44.57) +- (0.0, 3.21)
(1200, 47.93) +- (0.0, 1.89)

    };\addlegendentry{CLINN-jr1}
    \addplot+[color=brown,mark=pentagon*,mark options={brown},error bars/error bar style={brown},error bars/.cd, y dir=both,y explicit]
    coordinates {
    (150, 42.97) +- (0.0, 2.2)
(300, 44.45) +- (0.0, 1.97)
(450, 44.73) +- (0.0, 2.76)
(1200, 47.75) +- (0.0, 1.8)
    };\addlegendentry{CLINN-jr2}
    \addplot+[color=green,mark=pentagon*,mark options={green},error bars/error bar style={green},error bars/.cd, y dir=both,y explicit]
    coordinates {
    (150, 43.18) +- (0.0, 2.54)
(300, 44.72) +- (0.0, 2.66)
(450, 44.67) +- (0.0, 3.09)
(1200, 47.93) +- (0.0, 1.89)

    };\addlegendentry{CLINN-jr3}
    
    \nextgroupplot[title=Joint Goal]
    %medium joint goal
    \addplot+[error bars/.cd, y dir=both,y explicit]
    coordinates {
    (150, 45.8) +- (0.0, 6.27)
(300, 56.67) +- (0.0, 4.34)
(450, 60.8) +- (0.0, 3.81)
(1200, 65.88) +- (0.0, 4.23)

    };%\addlegendentry{DAMD}
    \addplot+[color=orange,mark=diamond*,mark options={orange},error bars/error bar style={orange},error bars/.cd, y dir=both,y explicit]
    coordinates {
    (150, 56.12) +- (0.0, 4.35)
(300, 64.13) +- (0.0, 1.98)
(450, 66.82) +- (0.0, 1.58)
(1200, 70.33) +- (0.0, 2.5)

    };%\addlegendentry{Expert1}
    \addplot+[color=red,mark=diamond*,mark options={red},error bars/error bar style={red},error bars/.cd, y dir=both,y explicit]
    coordinates {
    (150, 54.62) +- (0.0, 2.98)
(300, 62.5) +- (0.0, 1.95)
(450, 64.13) +- (0.0, 1.27)
(1200, 68.28) +- (0.0, 3.1)

    };%\addlegendentry{CLINN (expert2)}
    % \addplot+[error bars/.cd, y dir=both,y explicit]
    % coordinates {};%\addlegendentry{Intermediate}
    \addplot+[color=cyan,mark=triangle,mark options={cyan},error bars/error bar style={cyan},error bars/.cd, y dir=both,y explicit]
    coordinates {
    (150, 55.33) +- (0.0, 3.23)
(300, 61.63) +- (0.0, 1.42)
(450, 63.85) +- (0.0, 1.2)
(1200, 67.85) +- (0.0, 1.89)

    };%\addlegendentry{Junior1}
    \addplot+[color=brown,mark=pentagon*,mark options={brown},error bars/error bar style={brown},error bars/.cd, y dir=both,y explicit]
    coordinates {
    (150, 55.33) +- (0.0, 3.23)
(300, 61.63) +- (0.0, 1.42)
(450, 63.85) +- (0.0, 1.2)
(1200, 67.8) +- (0.0, 1.9)

    };%\addlegendentry{Junior2}
    \addplot+[color=green,mark=pentagon*,mark options={green},error bars/error bar style={green},error bars/.cd, y dir=both,y explicit]
    coordinates {
    (150, 55.33) +- (0.0, 3.23)
(300, 61.63) +- (0.0, 1.42)
(450, 63.85) +- (0.0, 1.2)
(1200, 67.85) +- (0.0, 1.89)

    };%\addlegendentry{CLINN (junior3)}
    
    \nextgroupplot[title=Slot Acc.]
    %medium joint goal
    \addplot+[error bars/.cd, y dir=both,y explicit]
    coordinates {
(150, 95.8) +- (0.0, 0.9)
(300, 97.17) +- (0.0, 0.48)
(450, 97.48) +- (0.0, 0.23)
(1200, 98.07) +- (0.0, 0.25)

    };%\addlegendentry{DAMD}
    \addplot+[color=orange,mark=diamond*,mark options={orange},error bars/error bar style={orange},error bars/.cd, y dir=both,y explicit]
    coordinates {
(150, 96.7) +- (0.0, 0.8)
(300, 97.68) +- (0.0, 0.24)
(450, 97.92) +- (0.0, 0.19)
(1200, 98.33) +- (0.0, 0.18)

    };%\addlegendentry{Expert1}
    \addplot+[color=red,mark=diamond*,mark options={red},error bars/error bar style={red},error bars/.cd, y dir=both,y explicit]
    coordinates {
    (150, 96.65) +- (0.0, 0.72)
(300, 97.62) +- (0.0, 0.25)
(450, 97.77) +- (0.0, 0.16)
(1200, 98.2) +- (0.0, 0.19)

    };%\addlegendentry{CLINN (expert2)}
    % \addplot+[error bars/.cd, y dir=both,y explicit]
    % coordinates {};%\addlegendentry{Intermediate}
    \addplot+[color=cyan,mark=triangle,mark options={cyan},error bars/error bar style={cyan},error bars/.cd, y dir=both,y explicit]
    coordinates {
    (150, 96.73) +- (0.0, 0.67)
(300, 97.55) +- (0.0, 0.26)
(450, 97.7) +- (0.0, 0.12)
(1200, 98.17) +- (0.0, 0.15)

    };%\addlegendentry{Junior1}
    \addplot+[color=brown,mark=pentagon*,mark options={brown},error bars/error bar style={brown},error bars/.cd, y dir=both,y explicit]
    coordinates {
    (150, 96.73) +- (0.0, 0.67)
(300, 97.55) +- (0.0, 0.26)
(450, 97.7) +- (0.0, 0.12)
(1200, 98.17) +- (0.0, 0.15)

    };%\addlegendentry{Junior2}
    \addplot+[color=green,mark=pentagon*,mark options={green},error bars/error bar style={green},error bars/.cd, y dir=both,y explicit]
    coordinates {
    (150, 96.73) +- (0.0, 0.67)
(300, 97.55) +- (0.0, 0.26)
(450, 97.7) +- (0.0, 0.12)
(1200, 98.17) +- (0.0, 0.15)

    };%\addlegendentry{CLINN (junior3)}
    
    \nextgroupplot[title=Slot F1]
    %medium joint goal
    \addplot+[error bars/.cd, y dir=both,y explicit]
    coordinates {
    (150, 80.33) +- (0.0, 4.98)
(300, 87.6) +- (0.0, 2.08)
(450, 88.87) +- (0.0, 1.05)
(1200, 91.5) +- (0.0, 1.1)

    };%\addlegendentry{DAMD}
    \addplot+[color=orange,mark=diamond*,mark options={orange},error bars/error bar style={orange},error bars/.cd, y dir=both,y explicit]
    coordinates {
    (150, 84.88) +- (0.0, 4.14)
(300, 89.83) +- (0.0, 1)
(450, 90.75) +- (0.0, 0.78)
(1200, 92.67) +- (0.0, 0.83)

    };%\addlegendentry{Expert1}
    \addplot+[color=red,mark=diamond*,mark options={red},error bars/error bar style={red},error bars/.cd, y dir=both,y explicit]
    coordinates {
    (150, 84.72) +- (0.0, 3.74)
(300, 89.63) +- (0.0, 1.16)
(450, 90.18) +- (0.0, 0.81)
(1200, 92.22) +- (0.0, 0.86)

    };%\addlegendentry{CLINN (expert2)}
    % \addplot+[error bars/.cd, y dir=both,y explicit]
    % coordinates {};%\addlegendentry{Intermediate}
    \addplot+[color=cyan,mark=triangle,mark options={cyan},error bars/error bar style={cyan},error bars/.cd, y dir=both,y explicit]
    coordinates {
    (150, 85.28) +- (0.0, 3.56)
(300, 89.32) +- (0.0, 1.11)
(450, 89.98) +- (0.0, 0.61)
(1200, 92) +- (0.0, 0.65)

    };%\addlegendentry{Junior1}
    \addplot+[color=brown,mark=pentagon*,mark options={brown},error bars/error bar style={brown},error bars/.cd, y dir=both,y explicit]
    coordinates {
    (150, 85.28) +- (0.0, 3.56)
(300, 89.3) +- (0.0, 1.08)
(450, 89.97) +- (0.0, 0.6)
(1200, 91.98) +- (0.0, 0.66)

    };%\addlegendentry{Junior2}
    \addplot+[color=green,mark=pentagon*,mark options={green},error bars/error bar style={green},error bars/.cd, y dir=both,y explicit]
    coordinates {
    (150, 85.28) +- (0.0, 3.56)
(300, 89.3) +- (0.0, 1.08)
(450, 89.98) +- (0.0, 0.61)
(1200, 92) +- (0.0, 0.65)

    };%\addlegendentry{CLINN (junior3)}
    
     \end{groupplot}
    \node (title) at ($(group c1r1.north)!0.5!(group c4r1.north)+(0,1cm)$) {\textbf{Free Rule Type - Medium Set}};
    %\path (top)--(bot) coordinate[midway] (group center);
    
    \node[anchor=north] (title-x) at ($(group c1r1.south east)!0.5!(group c4r1.south west)-(0,0.5cm)$) {Training Examples};
    
    \node[anchor=north] (title-x) at ($(group c1r1.south east)!0.5!(group c4r1.south west)-(0,1cm)$){\pgfplotslegendfromname{models}};

    \end{tikzpicture}

    \caption{Trend of models' performances when increasing the number of dialogue examples used to train DAMD. The plots are showing results of the configuration where CLINN uses the \textit{Medium} set of the \textit{Free} Rule type}
    \label{fig:medium-free-rule-metrics}
\end{figure*}

\subsection{Results and Discussion}
\label{sec:discussion}

Results of the experiments are extremely interesting both for industrial practice and for research. In this section, we analyze these results. Firstly, we analyze the relative quality \completare{and agreement level} of the conversational designers. Secondly, we investigate the quality of the produced rule sets in CLINN. Finally, we describe the limitations of our study. 

%The different expertise of conversational designers is immediately clear (see Table \ref{tab:Inter-annotator}).
\completare{
Our first observation is that it is difficult to write rules and agreement in writing these rules seems to decrease with experience (see Table \ref{tab:Inter-annotator}). 
%we have the agreement scores between pairs of conversational designers, which are used to understand how the difference in skills is reflected on the rule production.
The Small set configuration of rules does not show a remarkable difference in agreement between expertise categories, perhaps due to the low number of rules produced for this set size. We notice that jr1 has low agreement scores with others for belief rules ($\leq 0.53$) with respect to other scores, while jr3 has the highest agreements with exp2 ($0.76$) and jr2 ($0.77$). Moreover, exp1 has low agreement with all annotators ($\leq 0.53$). In the case of action rules, the relevant agreements are achieved by jr3 with exp1 ($0.41$) and jr2 ($0.66$), while other scores are less than $0.38$ clearly showing a type of rules that is subjected to designer's experience. These agreement scores are reflected in results, as performances are very similar between designers when using this set of rules, except for jr1 whose performances are slightly better than the others' in the Free rule configuration.
}
\completare{
A bigger difference in agreement is shown for the medium rules sets, probably due to the increased size of dialogue examples. There is an evident difference between expertise classes for both belief and action rules. In particular, high levels of agreement are detected between juniors' rules for both belief ($\geq 0.84$) and action ($\geq 0.4$) rules with respect to other pairs. Moreover, for action rules we have that agreement scores between classes are lower than $0.4$, which may be a signal of a complex kind of rules that is prone to subjectivity during the design process. The fact that, for belief rules, the highest agreement scores are between juniors provide additional support to our previous observation that experience leads towards a different view on dialogue systems design, hence obtaining lower accordance among designers. Results obtained from experts significantly outperforms the other ones in all metrics for this size of rules set, although the difference is minimal when considering the Action F1 measure.}
\completare{From this analysis, it is reasonable to assume that a high agreement between rules designed by juniors could be a consequence of less experience that prevents a subjective interpretation of the dialogues.
Juniors mainly rely on given instructions. Hence, they obtain similar sets of rules. Moreover, in the case of belief rules, there is a moderate ($>0.5$), or high, overall agreement for the majority of designers pairs. This suggests that the nature of these rules reduces the space of subjective interpretation.}

The next step is to analyze the performance of dialogue systems using the transition rule sets produced by the different designers. Results are reported in Table \ref{tab:pact-results}. To compare results, we report results of two configurations of the fully neural-based dialogue system: DAMD, which is our baseline, and Fully-informed DAMD, which is our upper-bound. These two configurations are useful to understand if transition rules are effective or not. For example, there is a very small space of improvement in measures like Action F1 -- $1.05$ difference in mean -- and Slot Accuracy -- $1.20$ difference in mean. Moreover, the results in all metrics where Fully-informed DAMD surpasses DAMD are statistically significative. 
%Moreover, the results in term Action F1 and Slot Accuracy of the two systems  -- DAMD and Fully-informed DAMD -- are not statistically different.  

It seems to be proven that only transition rules produced by designers are not sufficient to build up a dialogue system that can efficiently handle testing dialogues.
%mic: cambierei "statisfactory" in "good enough"
Results of CLINN base are not statisfactory. There is not any configuration that is in between the baseline (DAMD) and the upper bound. This suggests that an integration between rule-based dialogue systems and neural-base dialogue systems is desired. \completare{This observation is also supported by the fact that using jr1's small rule set with DAMD results in performances similar or higher than the ones achieved with rules provided by other juniors, despite its very low measures obtained in CLINN base mode.}

The integration of designers' rules in the overall system is effective and useful. In fact, all the CLINN configurations outperform the DAMD system in Joint Goal, Slot Accuracy, and Slof F1 metrics. The difference, except for some cases, is statistically significant. \completare{In the case of Action F1, CLINN significantly outperforms DAMD for the majority of the Medium rule set configurations, while Small rule set seem not very effective for this measure according to the lack of significant improvements in performances. Moreover, the rules designed by experts achieve overall better scores than other configurations, while juniors' ones have statistically significant better scores only for Free rule type. For these configurations exp1's rule set achieves the best scores in all the metrics.} %In the case of Action F1, there are only two configurations of CLINN that significantly outperform DAMD, that is, CLINN Medium set, Free rules produced by junior1 and junior2.

Using transition rules with less constraints seems to be the way to go when used in combination with DAMD. The configuration \emph{Free} is better than the configuration \emph{Full} for \completare{quite} all the measures in both the Small and the Medium settings.
\completare{The most noticeable improvements in performances are found in the Joint Goal metric: CLINN achieves improvements of at least $3.4$ difference in mean in the case of Small rule sets, while for Medium rule sets the performances gain at least $2.9$ of improvement. The highest gains in scores are obtained by using juniors' Medium rule sets of $4.55$ difference in mean.
The same observation applies for CLINN base, where the same behaviour is observed for all metrics including Action F1. Hence,} Over-constraining rules seems to be not effective \completare{when applied with DAMD}.

Experience in writing transition rules is important. \completare{In fact, for all measures involving the belief, expert's rules are behaving better than juniors' rules.} %Moreover, even a combination of rules from junior1 and junior2 (CLINN-merge) is not more effective than rules for the expert. 
\completare{We noted that rules provided by experts significantly outperform DAMD in all metrics for all configurations, except in the case of Action F1 metric where for Small rule sets results do not significantly improve with respect to the ones obtained by DAMD alone.}
Moreover, the experts are able to gain more effective rules by reading additional dialogues. The difference between Small-Free and Medium-Free is higher for the experts than for the juniors in the case of Joint Goal.

Finally, using experts to build up transition rules seems to be better than using effort in annotating additional dialogues. In fact, adding training examples to DAMD does not clearly outperform CLINN with expert rules (Figure \ref{fig:medium-free-rule-metrics}). DAMD with 1,200 examples behaves similarly to CLINN with expert's rules that uses 450 training examples. This latter is even close to CLINN with expert's rules using only 300 training examples. 
This is a very important observation as suggests a clear view for where to invest time and efforts.

There are, of course, some limitations of this study to acknowledge.
Firstly, actions produced by DAMD and CLINN do not contain values of informed slots; this prevents belief state trackers from accessing to additional information that should be tracked in the next turn.
Secondly, when CLINN produces only the belief state $B_t$, the action $A_t$ generated by DAMD, which will be forwarded to the next turn, is still generated according to DAMD's $B_t$; this is due to the architecture of DAMD that prevents the replacement of the hidden representation of DAMD's $B_t$ with the symbolic CLINN's $B_t$. However, these limitations do not falsify our previous conclusions.

%\meta{about these limitations \dots}
%\meta{explain why problems not addressed?}
%for DST (delex sys acts): we are only considering a setting where the user explicitly inform the system about values. it's a simplified scenario (but still considering wrong restaurant name prediction as error during evaluation, although it will be ever wrong without info from the system)

\section{Conclusions}
Merging pre-existing explicit knowledge and learning from examples is one of the most important research line in studies in learning neural networks and, in general, in machine learning.
Yet, there is not a clear understanding on how the quality of the teachers affects results of final systems.  

In this paper, we carried out a study on how rules provided by conversational designers affect the performance of neural-based dialogue systems. %In this paper, we carried out a study on how the experience of conversational designers affects the performance of neural-based dialogue systems. 
\completare{We firstly collected different sets of rules derived from task-oriented dialogue systems implemented by different-skilled conversational designers, then we combined them with a neural-based dialogue system by applying these rules to situations in dialogue for which they are appropriate.}
Our results are an important indication as we showed that designers can significantly reduce the sets of annotated dialogue examples, especially in the case of more experienced designers. %Our results are an important indication as we showed that experienced designers can significantly reduce the sets of annotated dialogue examples. 
\completare{Moreover, we gained some insights about how different skills of designers affect dialogue systems designing and, hence, their performances.}

Therefore, as a general contribution, our study showed that\completare{, in contrast with the main stream in natural language processing.} companies developing dialog systems should invest more in experienced conversational designers and less in extensive dialogue collection and annotation.  
\bibliographystyle{acl_natbib}
\bibliography{references}

%\appendix

\end{document}